\documentclass[
]{ceurart}

\usepackage{listings}
\usepackage{graphicx}
\usepackage{array}   
\usepackage{amsfonts}
\usepackage{amsmath}
\usepackage{amssymb}
\usepackage{algorithm}
\usepackage{algpseudocode}
\usepackage{xcolor}
\usepackage{booktabs}
\usepackage{tabularx}
\usepackage{multicol}
\usepackage{multirow}
\usepackage{pifont}
\usepackage{mdframed}
\usepackage{spverbatim}
\usepackage{comment}
\usepackage{enumitem}
\usepackage{hyperref}

\newcommand{\cmark}{\ding{51}}
\newcommand{\xmark}{\ding{55}}%
\begin{document}

\copyrightyear{2026}
\copyrightclause{Copyright for this paper by its authors. Use permitted under Creative Commons License Attribution 4.0 International (CC BY 4.0).}

\conference{ISWC 2026 Companion Volume, October 25--29, 2026, Bari, Italy}

\title{APOLO: Automatic Prompt Optimization for Ontology Learning}


\author[1,3]{Huu Tan Mai}[%
email=huutan.mai@de.bosch.com,
]
\cormark[1]
\fnmark[1]

\author[1,4]{Roman Kochnev}[%
email=roman.kochnev@stud-mail.uni-wuerzburg.de,
]
\cormark[1]
\fnmark[1]

\author[1]{Cuong Xuan Chu}[%
]

\author[2]{Lukas Lange}[
]

\author[3]{Heiko Paulheim}[%
]

\author[1]{Daria Stepanova}[%
]

\address[1]{Bosch Corporate Research, Robert-Bosch-Campus 1, 71272 Renningen, Germany}
\address[2]{Bosch Research North America, Sunnyvale, California, USA}
\address[3]{University of Mannheim, Schloss, 68131 Mannheim, Germany}
\address[4]{University of Würzburg, Sanderring 2, 97070 Würzburg, Germany}

\cortext[1]{Corresponding author.}
\fntext[1]{These authors contributed equally.}

\begin{abstract}
    Ontology Learning (OL) from text has advanced with the emergence of Large Language Models (LLMs), but it remains challenging due to the limited availability of annotated training data and the difficulty of adapting LLMs to perform OL effectively. We address this via APOLO -- Automatic Prompt Optimization for Ontology Learning, by casting OL as an explicit prompt optimization problem over LLM modules. To obtain training data, we employ a multi-agent system that generates text-ontology pairs from existing expert-curated ontologies. We then propose two ontology learner architectures: a greedy and an autoregressive learner, and optimize both using GEPA, a greedy evolutionary prompt optimizer built on DSPy. Experiments on two ontologies, a biomedical (DOID) and a plant ontology (PO), show consistent improvements after optimization across nearly all model and mode combinations, with autoregressive learners achieving the largest gains. Our results demonstrate that prompt optimization is a viable and lightweight alternative to fine-tuning for OL, and that the autoregressive formulation better captures ontological structure than the greedy approach.
\end{abstract}


\maketitle

\section{Introduction}
\label{sec:introduction}

Ontologies are formal frameworks for representing knowledge, enabling trustworthy AI services, seamless integration of heterogeneous data, and symbolic reasoning~\cite{10.1145/3447772}. Yet constructing domain ontologies remains a costly, expertise-intensive process~\cite{buitelaar2005ontology}. Ontology Learning (OL) seeks to automatically extract formal knowledge from text, with the goal of reducing this burden. Recent advances in Large Language Models (LLMs) and agentic AI 
are enabling the automation of OL~\cite{giglou2023llms4ollargelanguagemodels,bakker2024ontology,rahnamoun2025sbu}. However, the search for the right setup to carry out the task optimally is time-consuming and resource-intensive. For instance, given a pipeline of agents that execute subtasks (e.g., systems~\cite{rahnamoun2025sbu} following the LLMs4OL paradigm~\cite{giglou2023llms4ollargelanguagemodels}), fine-tuning each LLM for its respective subtask is costly. Similarly, although it has been shown that the performance of LLMs on downstream tasks is highly sensitive to the choice of prompt~\cite{liu2023pre}, manually searching for a high-performance prompt is time-consuming. To address these challenges, automatic \emph{prompt optimization} offers a lightweight alternative: it improves model behavior without any weight updates, making it significantly cheaper than supervised fine-tuning or reinforcement learning while remaining applicable to any LLM.
Building on this motivation, we make the following contributions:
\begin{itemize}
    \item We formalize OL as an explicit prompt optimization problem with respect to LLM prompts over text-ontology pairs -- a framing that directly supports this approach.
    \item We propose two learner architectures: a greedy learner and an autoregressive (AR) learner -- both optimized using GEPA~\cite{agrawal2025gepa}, an evolutionary prompt optimizer built on DSPy~\cite{khattab2024dspy}.
    \item We empirically show that APOLO yields consistent improvements for OL across nearly all LLM and method combinations on a challenging OL benchmark.
\end{itemize}

\section{Approach}
\label{sec:approach}

\begin{figure}[t]
    \centering
    \includegraphics[width=\linewidth]{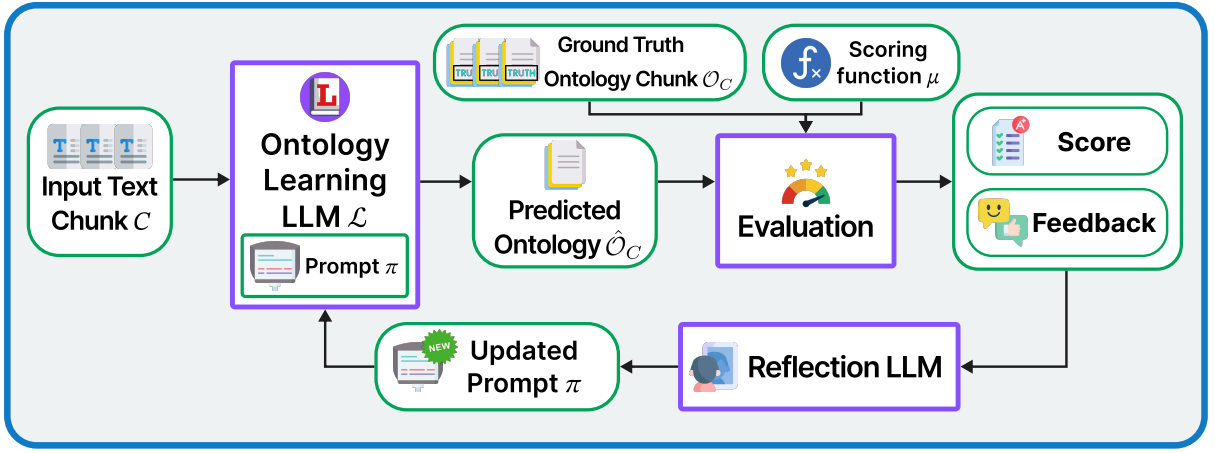}
    \caption{
    Overview of APOLO's prompt optimization framework. An ontology learner predicts an ontology from an input text chunk using prompt $\pi$. The prediction is evaluated against a ground-truth ontology chunk to produce a score and actionable feedback; a reflection LLM uses both to iteratively improve $\pi$ while keeping the underlying LLM fixed.}
    \label{fig:prompt_optimization_pipeline}
\end{figure}

\paragraph{Ontology Learning as Optimization.}
An \emph{ontology} formally represents domain concepts and their relationships as a collection of \emph{axioms} over classes, properties, and individuals in 
Description Logics~\cite{baader2003description,krotzsch2012description}.
We assume 
access to document-ontology pairs $(D,\mathcal{O})$, where $D$ is an input document, and $\mathcal{O}$ its ground-truth ontology, drawn from a distribution $\mathcal{P}$ (e.g., where $D$ is generated by an LLM from $\mathcal{O}$). 
An 
ontology learner $\mathcal{L}$ 
maps a document $D$ to a predicted ontology $\hat{\mathcal{O}} = \mathcal{L}(D)$. Given an evaluation metric $\mu(\hat{\mathcal{O}},\mathcal{O})$ that 
scores a prediction against ground truth, 
casting OL as an optimization problem 
amounts to optimizing $\max_{\mathcal{L}}\ \mathbb{E}_{(D,\,\mathcal{O})\sim\mathcal{P}}
    \left[\mu(\mathcal{L}(D),\,
    \mathcal{O})\right]$.
Alternatively, 
a document $D$ and its ontology $\mathcal{O}$ can be 
split into corresponding text chunks $C\in \mathrm{chunks}(D)$, each paired with the subontology $\mathcal{O}_C$ grounded in $C$. 
When inference must operate on chunks (e.g., due to context length limits), 
the learner predicts over a single chunk $C$ in the context of $D$. 
Concretely, given multiple pairs $(D^{(i)}, \mathcal{O}^{(i)})$, we 
form a chunk-level dataset by pairing every chunk with its parent document and the subontology it grounds, 
$\mathcal{T} = \bigl\{\, (C,\, D^{(i)},\, \mathcal{O}^{(i)}_C) \;:\; C \in \mathrm{chunks}(D^{(i)}) \,\bigr\},$
where $\mathcal{O}^{(i)}_{C}\subseteq \mathcal{O}^{(i)}$ denotes a subontology of $\mathcal{O}^{(i)}$ grounded in chunk $C$.
We then optimize the following objective
\begin{equation}\label{eq:surrogate-method}
    \max_{\mathcal{L}} \; \mathbb{E}_{(C, D, \mathcal{O}_C) \sim \mathcal{T}} \left[ \mu \left( \mathcal{L}_{\pi}(C, D), \mathcal{O}_C \right) \right].
\end{equation}

\paragraph{Prompt Optimization for OL.} We now fix the learner's prompt $\pi$ as the only tunable parameter. More specifically, let $\mathcal{L}_\pi$ denote the learner parametrized by its prompt $\pi$ with the underlying weights kept fixed. \emph{Prompt optimization} searches for a prompt $\pi^{\star}$ that maximizes the expected score
\begin{equation}\label{eq:prompt-optimization}
\pi^{\star} =  \arg\max_{\pi} \;
    \mathbb{E}_{(C,\,D,\mathcal{O}_C)\,\sim\,\mathcal{T}}
    \left[\mu\!\left(\mathcal{L}_\pi(C,D),\, \mathcal{O}_C\right)\right],
\end{equation}
i.e., the empirical objective of Eq.~\ref{eq:surrogate-method} restricted to prompt-level parameters.
GEPA~\cite{agrawal2025gepa} is one such optimizer: it iteratively mutates prompt candidates and uses a separate LLM to reflect on failure cases and propose improvements. Crucially, prompt optimization leaves model weights unchanged, making it a lightweight alternative to fine-tuning or reinforcement learning with a limited budget of rollouts. Prior work applied this paradigm to knowledge graph construction~\cite{mihindukulasooriya2025automatic} using different prompt optimizers; we extend it to the more general setting of extracting ontological elements, and in particular, classes, individuals and OWL 2 axioms from text. 
This framework, which forms the core of APOLO, is directly applicable to OL over text chunks.
Figure~\ref{fig:prompt_optimization_pipeline} illustrates the proposed application of the GEPA prompt optimization loop to OL. Given an input text, the learner predicts an ontology using the current prompt $\pi$, which is manually initialized. The prediction is evaluated against the ground-truth ontology to produce both a quality score and actionable feedback, which are used by a reflection LLM to iteratively improve the prompt. Next, we define 
(a) an ontology learning architecture to optimize w.r.t. a prompt, (b) an evaluation metric $\mu$, and (c) actionable feedback in the context of GEPA. 

\begin{figure}[t]
    \centering
    \includegraphics[width=0.99\linewidth]{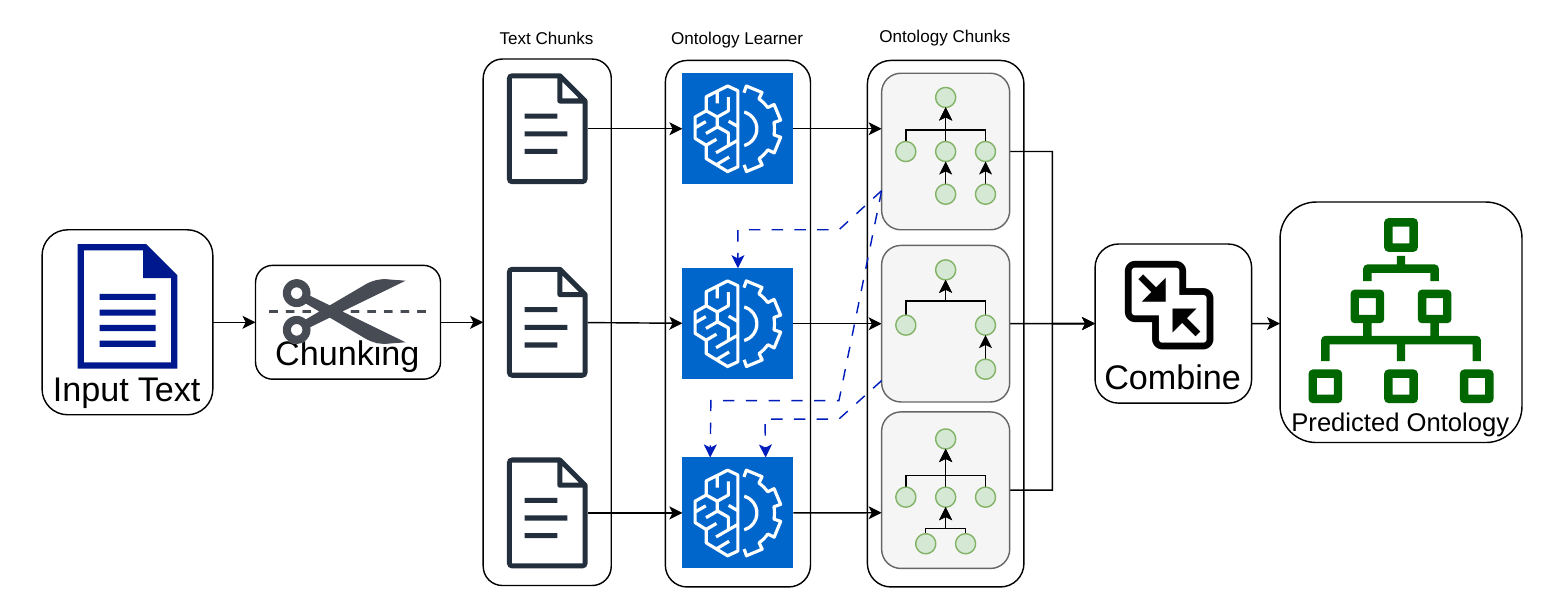}
    \caption{Overview of Greedy and Autoregressive learners at inference time. The \textbf{Greedy learner} splits documents into manageable chunks, each of which is turned into a corresponding ontology chunk. Once all chunks are constructed, they are concatenated into the resulting ontology. The \textbf{Autoregressive learner} additionally takes prior ontology chunks as context when generating a new chunk (dashed blue lines \textcolor{blue}{$\dashrightarrow$}).}
    \label{fig:ol-methods}
\end{figure}

\paragraph{Ontology Learning LLM.}
We propose two single-module (and therefore single-prompt) APOLO learners, whose prompts can be optimized with GEPA~\cite{agrawal2025gepa}. An overview is presented in Figure~\ref{fig:ol-methods}.

\emph{Greedy Learner.} This learner consists of a chunk-level base learner $\mathcal{L}_\pi^{\mathrm{gbase}}$ that processes each chunk $C$ 
separately conditioned on 
the full text $D$, and the 
resulting ontologies are 
concatenated. In essence, this assumes that the global solution can be reached by combining locally optimal ones.
We optimize the chunk-level surrogate objective (as per Equation~\ref{eq:surrogate-method}) w.r.t. the prompt $\pi$ of $\mathcal{L}_{\pi}^{\mathrm{gbase}}$ (see Equation~\ref{eq:prompt-optimization}). 

\emph{Autoregressive Learner.} 
This learner likewise constructs an ontology from chunk-level predictions, but its base learner 
additionally conditions each chunk on axioms predicted for preceding chunks, maintaining continuity across the document. During optimization, \emph{teacher forcing} is applied: ground-truth axioms from preceding chunks serve as context instead of the model's own predictions to prevent error accumulation.

\paragraph{Evaluation.} While in the presented framework any evaluation function $\mu$ for comparing the predicted ontology with the ground truth can be plugged in, we instantiate $\mu$ as follows. 
Entity precision (EnP) and recall (EnR) measure how well classes and individuals are recovered. Axiom precision (AxP) and recall (AxR) use entailment-based matching computed with the help of a reasoner, and measure the correctness, i.e., predicted axioms that are entailed by $\mathcal{O}$, and the completeness, i.e., ground-truth axioms that are entailed by $\hat{\mathcal{O}}$, of the prediction.
The scoring function with hyperparameters $\beta\in[0,\infty)$ and $\delta\in(0,1)$ is defined as follows:
\begin{align}
    \mu(\hat{\mathcal{O}}, \mathcal{O}) =
    \begin{cases}
        0 & \text{if } \hat{\mathcal{O}}
            \text{ is inconsistent or unparsable,}\\
        \delta\cdot\mathrm{EnF}(\beta) +
        (1\!-\!\delta)\cdot\mathrm{AxF}(\beta)
            & \text{otherwise,}
    \end{cases}
\end{align}
where $\mathrm{EnF}(\beta)$ and $\mathrm{AxF}(\beta)$ are the F-$\beta$ scores induced by the respective precision/recall pairs. 
Feedback should include error diagnostics such as: (a) explanations if $\hat{\mathcal{O}}$ is inconsistent or unparsable; (b) axioms from $\mathcal{O}$ not entailed by $\hat{\mathcal{O}}$; (c) axioms from $\hat{\mathcal{O}}$ not entailed by $\mathcal{O}$; (d) entities from $\mathcal{O}$ missing from $\hat{\mathcal{O}}$.

\section{Experiments}
\label{sec:experiments}

\begin{figure}[t]
    \centering
    \includegraphics[width=\linewidth]{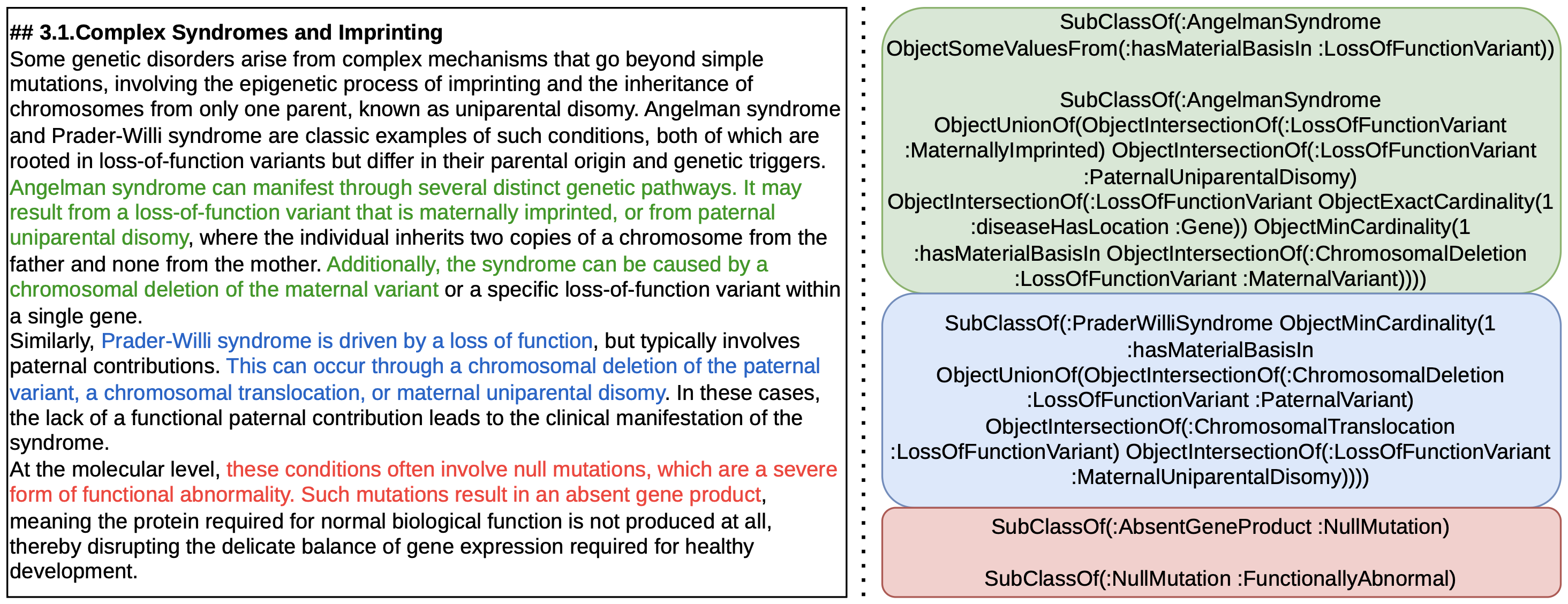}
    \caption{Example of a text-ontology training pair used for prompt optimization. 
    A text chunk generated from a subontology of the initial domain ontology is annotated with corresponding entities and axioms, each grounded in a supporting quote from the text.} 
    \label{fig:generated_text_axiom_pairs}
\end{figure}

\paragraph{Setup.}
A key obstacle for LLM-based OL from text is the near-absence of text-ontology pairs in which sentences are grounded in formal OWL axioms. Thus, we use
a multi-agent system to generate annotated text-ontology pairs from 12 publicly available ontologies spanning diverse domains (biomedical, e-commerce, chemistry, building services). The system comprises a Planner and Verifier for ontology decomposition and consistency checking, and a Writer and Critic for 
the generation of textual snippets from ontology axioms and their refinement. Figure~\ref{fig:generated_text_axiom_pairs} shows an example of a generated text-ontology pair. Two held-out ontologies serve as the common testbed: DOID~\cite{schriml2012disease} (human diseases, 1,407 words, 45 atomic axioms) and PO~\cite{avraham2008plant} (plant structure, 2,384 words, 53 atomic, 10 complex axioms). 
The remaining 10 ontologies, spanning different domains, are used for prompt optimization. GEPA optimization uses gpt-5-mini as the reflection LLM. We evaluate two prominent open-source LLMs as representatives: Qwen3.5-27B~\cite{qwen3.5} and Devstral-Small-2-24B~\cite{rastogi2025devstral}. We 
set $\beta=3$ and $\delta=0.2$ for the scoring function used during optimization.

\paragraph{Results.}
Table~\ref{tab:training-results-combined} shows that established OL baselines struggle on our benchmark, while APOLO's prompt-optimized learners achieve consistent improvements. Most baselines result in near-zero axiom-level scores, with one timing out. Bakker \emph{et al.}\ (B) -- the strongest baseline on DOID (AxP\,=\,0.667) -- collapses on PO (AxP\,=\,0.018), revealing a lack of cross-domain generalization. For our learners, two findings stand out: \emph{(i)} Improvements after GEPA optimization are consistent across nearly all model/mode combinations, suggesting that the generated data provides a reliable training signal. \emph{(ii)} APOLO's AR learners benefit most: Qwen3.5-27B AR 
achieves a 47\% improvement in AxP on DOID after optimization (i.e., from AxP\,=\,0.569 to AxP\,=\,0.838), and surpasses all off-the-shelf baselines across most axiom-level metrics.
A qualitative inspection reveals that optimized prompts typically become very verbose (e.g., from 14 words to 1474 words for Qwen3.5-27B AR), thoroughly detailing common patterns, formatting rules, quality goals, successful strategies, and pitfalls to avoid for ontology extraction. Unlike weight-update approaches, prompt optimization is compute-friendly: hosting one model with $M$ optimized prompts is substantially cheaper than $M$ fine-tuned models. In our experiments, GEPA optimization completed in approximately 3--4 hours with no GPU memory required for weight updates. Furthermore, GEPA is known to achieve large quality gains from few rollouts~\cite{agrawal2025gepa}, making it especially practical under limited compute budgets.

\begin{table}[t]
    \centering
    \caption{Comparison of baseline OL methods and prompt-optimized learners (Ours) on DOID and PO. \cmark/\xmark~denote after/before GEPA optimization; improvements over unoptimized in bold. Results are from a single run.}
    \label{tab:training-results-combined}
    \footnotesize
    \resizebox{\textwidth}{!}{
    \begin{tabular}{llccccccccccc}
    \toprule
       \multirow{2}*{Model} & \multirow{2}*{Method} & 
       \multirow{2}*{Opt.} & &
       \multicolumn{4}{c}{DOID} & & \multicolumn{4}{c}{PO} \\
       \cline{5-8} \cline{10-13}
       & & & & AxP & AxR & EnP & EnR & & AxP & AxR & EnP & EnR \\
    \midrule

    \multirow{9}*{Qwen3.5-27B}
      & AutoRAGLearner~\cite{ontolearner}
        & N/A & & .006 & .044 & .026 & .071 & & .006 & .048 & .057 & .141 \\
      & SBU-NLP~\cite{rahnamoun2025sbu}
        & N/A & & .000 & .000 & .074 & .143 & & .023 & .063 & .155 & .324 \\
      & Kommineni \emph{et al.}~\cite{kommineni2024human}
        & N/A & & .064 & .111 & .552 & .381 & & .000 & .000 & .189 & .099 \\
      & Bakker \emph{et al.}~(A)~\cite{bakker2024ontology}
        & N/A & & .583 & .622 & .725 & .690 & & .027 & .048 & .322 & .282 \\
      & Bakker \emph{et al.}~(B)~\cite{bakker2024ontology}
        & N/A & & .667 & .578 & .675 & .643 & & .018 & .048 & .180 & .282 \\
    \cline{2-13}
      & \multirow{2}*{Greedy (Ours)}
        & \xmark & & .457 & .689 & .569 & .690 & & .063 & .063 & .337 & .493 \\
      & & \cmark  & & \textbf{.585} & .689 & \textbf{.580} & .690
                   & & \textbf{.071} & \textbf{.111} & .281 & \textbf{.507} \\
    \cline{2-13}
      & \multirow{2}*{AR (Ours)}
        & \xmark & & .569 & .690 & .457 & .689 & & .092 & .079 & .362 & .479 \\
      & & \cmark  & & \textbf{.838} & .689 & \textbf{.829} & \textbf{.690}
                   & & \textbf{.174} & \textbf{.159} & \textbf{.425} & \textbf{.521} \\

    \midrule

    \multirow{9}*{Devstral-Small-2-24B}
      & AutoRAGLearner~\cite{ontolearner}
        & N/A & & .000 & .067 & .018 & .071 & & .005 & .063 & .022 & .085 \\
      & SBU-NLP~\cite{rahnamoun2025sbu}
        & N/A & & .037 & .000 & .102 & .143 & & \multicolumn{4}{c}{Timed out} \\
      & Kommineni \emph{et al.}~\cite{kommineni2024human}
        & N/A & & .209 & .267 & .579 & .524 & & .028 & .032 & .276 & .113 \\
      & Bakker \emph{et al.}~(A)~\cite{bakker2024ontology}
        & N/A & & .000 & .000 & .700 & .666 & & .000 & .000 & .000 & .000 \\
      & Bakker \emph{et al.}~(B)~\cite{bakker2024ontology}
        & N/A & & .000 & .000 & .286 & .667 & & .000 & .000 & .183 & .324 \\
    \cline{2-13}
      & \multirow{2}*{Greedy (Ours)}
        & \xmark & & .326 & .667 & .367 & .690 & & .035 & .079 & .174 & .338 \\
      & & \cmark  & & .284 & .622 & \textbf{.412} & .667
                   & & \textbf{.051} & .063 & \textbf{.218} & \textbf{.366} \\
    \cline{2-13}
      & \multirow{2}*{AR (Ours)}
        & \xmark & & .500 & .577 & .533 & .571 & & .009 & .016 & .244 & .422 \\
      & & \cmark  & & \textbf{.644} & \textbf{.622} & .528 & \textbf{.667}
                   & & \textbf{.094} & \textbf{.047} & \textbf{.326} & .408 \\

    \bottomrule
    \end{tabular}}
\end{table}

\paragraph{Limitations.} While we attempt to operationalize our evaluation as much as possible, our scoring method has limitations: it does not penalize redundant axioms -- that is, predicted axioms entailed by $\mathcal{O}$ are counted as correct regardless of redundancy; since optimization tends to produce verbose prompts, predicted ontologies may become overgenerated. Furthermore, to obtain ground-truth data for our text-ontology pairs, we use existing domain ontologies; however, these underlying ontologies may have been seen by the backbone LLMs during pretraining, which could inflate performance estimates.

\section{Conclusion}
\label{sec:conclusion}

We formalized Ontology Learning as an explicit prompt optimization problem over LLM modules, proposed two learner architectures -- greedy and autoregressive -- and optimized both with GEPA using generated text-ontology pairs. Experiments on two domain-specific benchmarks (DOID, PO) show that prompt optimization yields consistent improvements across nearly all model/mode combinations without any weight updates, with autoregressive learners benefiting most. Prompt-optimized Qwen3.5-27B AR surpasses all tested off-the-shelf baselines across most axiom-level metrics, demonstrating that even single-module APOLO learners can be meaningfully improved through this paradigm.

These results suggest that combining automatic data generation with prompt optimization is a promising path towards better ontology learners -- one that scales naturally to multi-agent architectures and arbitrary ontology domains. Future work will explore synthetic ontology generation to further reduce data leakage concerns, and more robust evaluation metrics for complex OWL 2 constructs.



\section*{Declaration on Generative AI}


\textbf{Claude (Anthropic)} was used exclusively for grammar 
and spelling checks, as well as minor stylistic improvements.
After using this tool, the authors reviewed and edited the paper as needed and take full responsibility for the 
content.

\bibliography{main}

\end{document}